\documentclass[10pt,twocolumn,letterpaper]{article}

\usepackage[pagenumbers]{cvpr} 

\usepackage{blindtext}
\usepackage{graphics}
\usepackage{caption}
\usepackage{colortbl}
\definecolor{mygray}{RGB}{234,234,234}

\definecolor{cvprblue}{rgb}{0.21,0.49,0.74}
\usepackage[pagebackref,breaklinks,colorlinks,citecolor=cvprblue]{hyperref}

\def\confYear{2025}

\title{EmoEdit: Evoking Emotions through Image Manipulation}

\begin{document}


\author{Jingyuan~Yang{\textsuperscript{1}},
	Jiawei~Feng{\textsuperscript{1}},
	Weibin~Luo{\textsuperscript{1}},
	Dani~Lischinski{\textsuperscript{2}},
	Daniel~Cohen-Or{\textsuperscript{3}},
	Hui~Huang{\textsuperscript{1}\thanks{Corresponding author}}\\
	\textsuperscript{1}CSSE, Shenzhen University 
	\textsuperscript{2}The Hebrew University of Jerusalem 
	\textsuperscript{3}Tel Aviv University \\	
	{\tt\small \{jingyuanyang.jyy, fengjiawei0909, waibunlok, danix3d, cohenor, hhzhiyan\}@gmail.com}
	\vspace{-25pt}
}

\twocolumn[{
	\renewcommand\twocolumn[1][]{#1}
	\maketitle
	\begin{center}
		\centering
		\includegraphics[width=\linewidth]{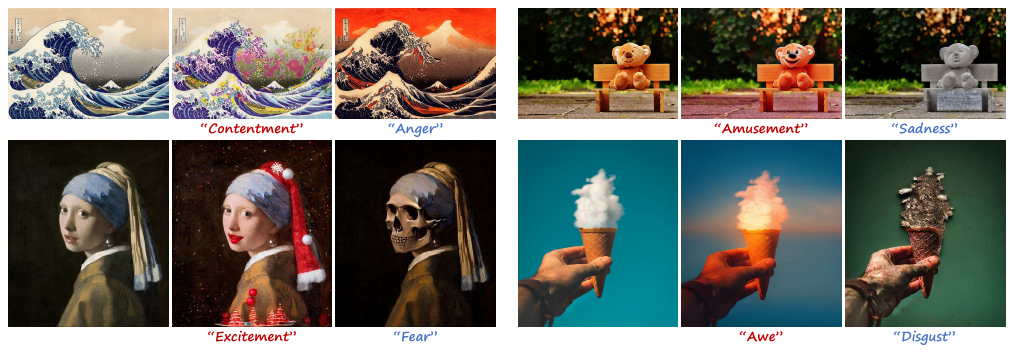}
		\captionof{figure}{Affective Image Manipulation with EmoEdit, which seeks to modify a user-provided image to evoke specific emotional responses in viewers. 
			Our method requires only emotion words as prompts, 
			without necessitating detailed descriptions of the input or output image.
		}
	\vspace{-3pt}
	\label{fig:teaser}
\end{center}
}]

\renewcommand{\thefootnote}{\fnsymbol{footnote}}
\footnotetext[1]{Corresponding author}

\begin{abstract}
	
Affective Image Manipulation (AIM) seeks to modify user-provided images to evoke specific emotional responses.
This task is inherently complex due to its twofold objective: significantly evoking the intended emotion, while preserving the original image composition.
Existing AIM methods primarily adjust color and style, often failing to elicit precise and profound emotional shifts.
Drawing on psychological insights, we introduce EmoEdit, which extends AIM by incorporating content modifications to enhance emotional impact.
Specifically, we first construct EmoEditSet, a large-scale AIM dataset comprising 40,120 paired data through emotion attribution and data construction.
To make existing generative models emotion-aware, we design the Emotion adapter and train it using EmoEditSet.
We further propose an instruction loss to capture the semantic variations in data pairs.
Our method is evaluated both qualitatively and quantitatively, demonstrating superior performance compared to existing state-of-the-art techniques.
Additionally, we showcase the portability of our Emotion adapter to other diffusion-based models, enhancing their emotion knowledge with diverse semantics.
	
\end{abstract}    
\section{Introduction}
\label{sec:intro}

\begin{flushleft}
	\textit{``The emotion expressed by wordless simplicity is the most abundant.''}
\end{flushleft}
\vspace{-18pt}
\begin{flushright}
	\textit{--William Shakespeare}
\end{flushright}
\vspace{-5pt}

Emotions weave through our daily lives, deeply intertwined with our perception of the world. Among myriad factors, visual stimuli stand out as particularly influential in shaping our emotional state. The emerging field of Visual Emotion Analysis (VEA)~\cite{rao2016learning,yang2018weakly,yang2023emoset}
explores the intricate connections between visuals and human emotions, with applications ranging from advertising~\cite{kang2020role,otamendi2020emotional} to robotics~\cite{gervasi2023applications,spitale2022affective}. In the arts, profound emotional responses have long been elicited via manipulation of various visual elements. This prompts an intriguing question: whether and how can we deliberately alter an image to steer the viewer's emotional experience?

This paper addresses the task of Affective Image Manipulation (AIM), which involves modifying a user-provided input image to evoke specific emotional responses in viewers, as depicted in Fig.~\ref{fig:teaser}. 
This task poses considerable challenges, requiring the automatic selection of appropriate visual elements suited for the image at hand, and integrating them in an emotionally resonant manner, while striving to maintain the original composition of the image. 

\begin{figure}
	\centering
	\includegraphics[width=\linewidth]{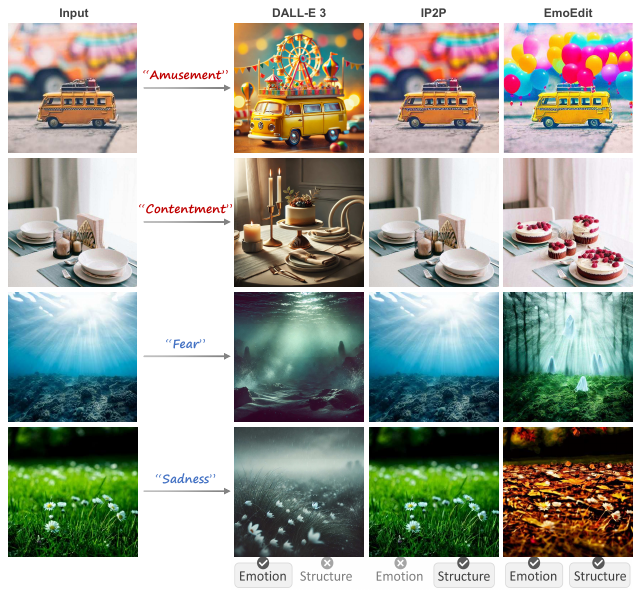}
	\vspace{-10pt}
	\caption{While DALL-E 3 conveys emotions well, IP2P remains faithful to original structure, neither approach satisfies both aspects. EmoEdit fills this gap by creating images with both emotion fidelity and structure preservation.
	}
	\label{fig:teaser_1}
	\vspace{-10pt}
\end{figure}

Current state-of-the-art generative models, while proficient in various image editing tasks, often fail to adequately balance AIM's contradictory objectives. For instance, DALL-E 3~\cite{betker2023improving} generates images that effectively convey the desired emotions but does not adhere to the original image structure, as shown in Fig.~\ref{fig:teaser_1}. Conversely, InstructPix2Pix (IP2P)~\cite{brooks2023instructpix2pix} remains faithful to the structure but lacks in emotional expressiveness. 

Previous approaches in AIM focused primarily on adjusting color and style \cite{wang2013affective, liu2018emotional, fu2022language, weng2023affective}, but these methods struggle to evoke precise and significant emotion shifts, and some are limited to binary emotion categories (positive and negative).
In this paper, we introduce EmoEdit, a novel approach inspired by psychological studies~\cite{brosch2010perception}, which goes beyond color and style adjustments to perform substantive modifications of relevant content.
Furthermore, we use the eight emotion categories introduced by Mikels \etal~\cite{mikels2005emotional}, to achieve more precise control over the emotions conveyed.

EmoEdit is a content-aware AIM framework capable of evoking emotions with diverse semantic modifications.
Given the lack of paired data, we first generate EmoEditSet, in two stages: emotion attribution and data construction.
Specifically, based on the recent large-scale EmoSet~\cite{yang2023emoset}, we leverage a Vision-Language Model (VLM) to create eight emotion factor trees, each comprising several semantic summaries for a specific emotion.
Images from several different sources~\cite{zhang2024magicbrush,shi2021learning} are collected and then modified by InstructPix2Pix~\cite{brooks2023instructpix2pix} with emotion factors to generate target candidates.
Since data quality is crucial for AIM, we carefully filter target candidates using four evaluation metrics and human feedback.
EmoEditSet ultimately comprises 40,120 image pairs, serving as a high-quality, semantic-diverse benchmark dataset for AIM.

To make diffusion models emotion-aware, we design the Emotion adapter, which integrates knowledge from EmoEditSet.
Specifically, we leverage the structure of Q-Former~\cite{li2023blip}, particularly its attention mechanisms, to facilitate interaction between the target emotion and the input image.  
To capture the semantic variations in emotion data pairs, we further introduce an instruction loss.
EmoEdit is optimized with both instruction loss and diffusion loss, yielding images that preserve the original structure while remaining emotionally faithful and semantically diverse.

To ensure fair comparisons, we assemble an inference set of 405 images with a distribution distinct from the training set.
Quantitative and qualitative evaluations are conducted against state-of-the-art editing techniques including: global, local, and emotion-related.
Our evaluation metrics assess three key aspects: pixel-wise similarity, semantic similarity, and faithfulness to the target emotion.
Finally, we demonstrate the portability of Emotion adapter to other diffusion-based models, spanning both editing and generation tasks.

In summary, our contributions are:
\begin{itemize}
	\setlength{\itemsep}{0pt}
	\setlength{\parsep}{0pt}
	\setlength{\parskip}{0pt}
	
	
	\item EmoEdit, a content-aware AIM framework capable of generating emotion-evoking, contextually fitting, and structurally faithful variant of a user-provided image, requiring only emotion words as prompts.
	
	\item EmoEditSet, the first large-scale AIM dataset, featuring 40,120 image pairs labeled with emotion directions and content instructions, establishing a high-quality, semantically diverse benchmark.
	
	\item Emotion adapter, trained with diffusion loss and the proposed instruction loss, functions as a plug-and-play module that enhances generative models with emotion-awareness once trained.  
	

\end{itemize}

\section{Related work}
\label{sec:rw}

\subsection{Visual Emotion Analysis}

Over the past two decades, researchers in VEA have focused on a pivotal inquiry: What evokes visual emotions? 
Responses have varied, ranging from low-level color and texture~\cite{lee2011fuzzy,machajdik2010affective,rao2020learning,zhang2019exploring} to high-level content and style~\cite{borth2013large,rao2020learning,zhang2019exploring,yang2021stimuli,yang2021solver}.
Lee \etal~\cite{lee2011fuzzy} propose an approach to evaluate visual emotions by constructing prototypical color images for each emotion.
Drawing inspiration from psychology and art theory, Machajdik \etal~\cite{machajdik2010affective} extract and combine color and texture features for emotion classification.
A significant milestone in this field was achieved by Borth \etal~\cite{borth2013large}, who constructed a comprehensive visual sentiment ontology named SentiBank, where each concept is represented by an Adjective-Noun Pair (ANP).
Rao \etal~\cite{rao2020learning} introduced MldrNet, a model that predicts emotions by combining pixel-level, aesthetic, and semantic features.
To develop a comprehensive representation for emotion recognition, Zhang \etal~\cite{zhang2019exploring} integrated content and style information.
Yang \etal~explored networks based on different visual stimuli \cite{yang2021stimuli}, and further investigated the correlations between them \cite{yang2021solver}.
While previous studies indicate a correlation between emotion and visual elements, establishing a causal relationship necessitates image manipulation.
By introducing specific visual elements into the input image and observing how the emotion varies, we can gain deeper insights into understanding emotions.

\subsection{Diffusion-based Image Manipulation}

Recent years have witnessed a meteoric rise in generative models, ranging from GANs~\cite{goodfellow2014generative}, VAEs~\cite{kingma2013auto}, and normalizing flows~\cite{rezende2015variational}, to diffusion models~\cite{ho2020denoising,rombach2022high,dhariwal2021diffusion}.
Methods in diffusion-based image manipulation can be roughly grouped into global~\cite{meng2021sdedit, tumanyan2023plug, wu2022unifying} and local~\cite{wallace2023edict, geng2023instructdiffusion, parmar2023zero, mokady2023null}.
A pioneering work, SDEdit~\cite{meng2021sdedit} leverages stochastic differential equations for guided image synthesis.
PnP~\cite{tumanyan2023plug} enables semantic editing through condition-controlled guidance.
To modify images in a more fine-grained manner, researchers have also explored local editing.
P2P-Zero~\cite{parmar2023zero} allows for zero-shot editing of images without training on paired data.
InsDiff~\cite{geng2023instructdiffusion} and IP2P~\cite{brooks2023instructpix2pix} allow users to edit specific regions of images by giving instructions.
ControlNet~\cite{zhang2023adding} enables precise, localized image editing through spatial conditioning in diffusion models.
Further, BlipDiff~\cite{li2024blip} integrates VLM with diffusion models for precise visual region modification.
By plugging an additional module, adapter-based methods~\cite{ye2023ip, mou2024t2i} successfully incorporate contextual information into diffusion models.
Existing editing methods can effectively manipulate concrete prompts (\ie, descriptions, instructions) but face challenges when dealing with more abstract emotions.
To address this, we design the Emotion adapter that enables diffusion-based models emotion-aware while preserving their original structure.

\subsection{Affective Image Manipulation}

Most of the previous works in AIM can be grouped into color-based \cite{yang2008automatic, wang2013affective, peng2015mixed, liu2018emotional, chen2020image, zhu2023emotional} and style-based \cite{fu2022language, sun2023msnet, weng2023affective}.
Yang \etal~\cite{yang2008automatic} pioneered the application of color transfer in image emotion manipulation, dividing the color spectrum into 24 distinct moods. 
Given an emotion word, Wang \etal~\cite{wang2013affective} automatically adjust image color to meet a desired emotion and Liu \etal~\cite{liu2018emotional} further consider the semantic information to prevent potential artifacts.
Peng \etal~\cite{peng2015mixed} propose a method to transfer the color and texture of the target image to the source image, to modify emotions.
Recent methods, such as CLVA~\cite{fu2022language} and AIF~\cite{weng2023affective} reflect emotions derived from textual input by adjusting the style of the original image.
Most previous studies have focused on adjusting color and style to elicit specific emotions.
However, psychologists have demonstrated that visual content is a critical emotional stimuli~\cite{brosch2010perception}.
In light of this, we introduce EmoEdit to evoke emotions through semantically diverse and contextual fitting modifications.
\section{Method}
\label{sec:method}

\begin{figure*}
	\centering
	\includegraphics[width=\linewidth]{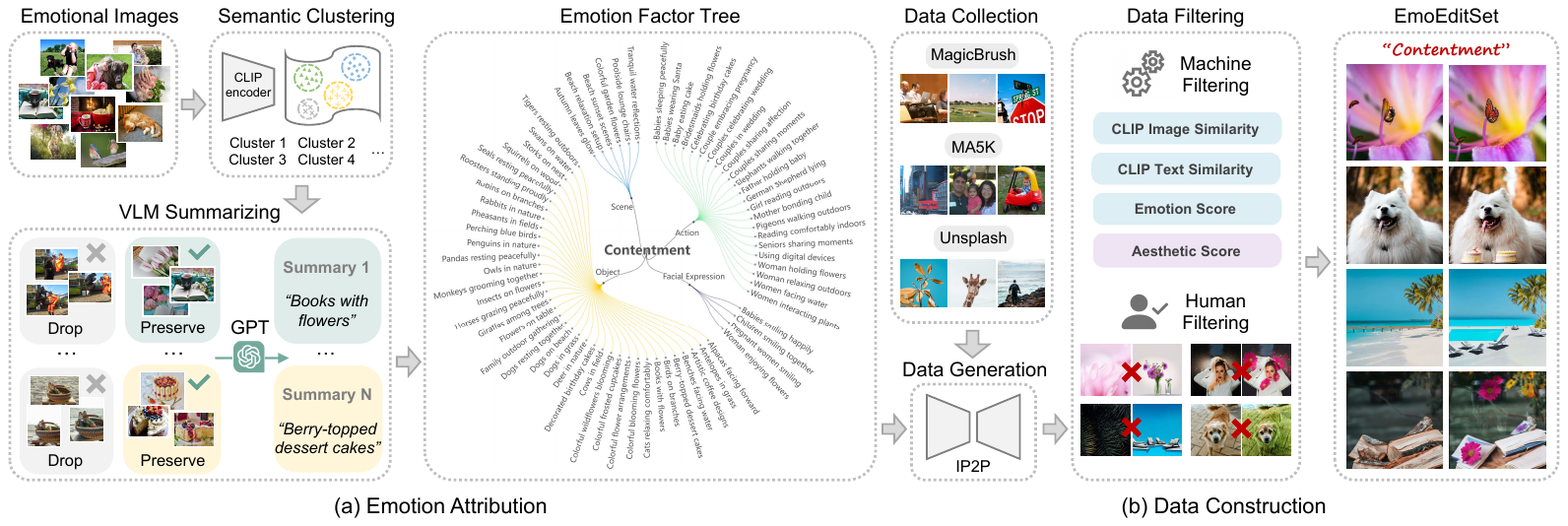}
	\vspace{-10pt}
	\caption{Overview of EmoEditSet. (a) Emotion factor trees are built with various representative semantic summaries based on EmoSet. (b) Through careful collection, generation and filtering, EmoEditSet is built with high-quality and semantic-diverse paired data.
	}
	\label{fig:method_1}
	\vspace{-10pt}
\end{figure*}

\begin{figure*}
	\centering
	\includegraphics[width=\linewidth]{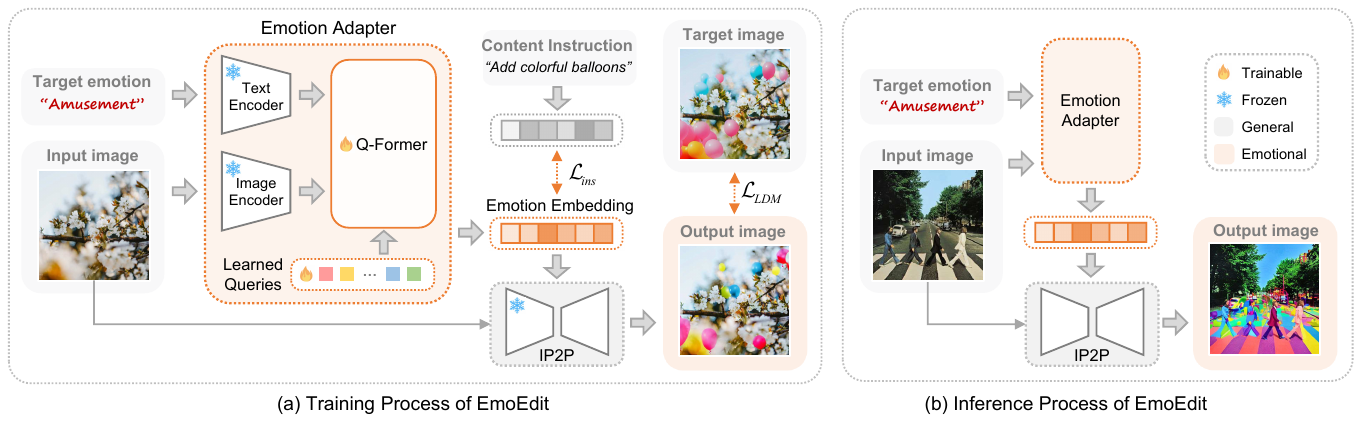}
	\vspace{-10pt}
	\caption{Overview of EmoEdit. (a) EmoEdit trains the Emotion adapter with paired data from EmoEditSet, by optimizing instruction loss and diffusion loss. (b) Given a user-provided image, EmoEdit can modify the image to evoke the desired emotion with clear semantics.
	}
	\label{fig:method_2}
	\vspace{-10pt}
\end{figure*}

We first construct EmoEditSet, the first large-scale AIM dataset (Fig.~\ref{fig:method_1}) and subsequently perform EmoEdit by training the Emotion adapter with the paired data (Fig.~\ref{fig:method_2}).

\subsection{EmoEditSet}

The absence of large-scale, high-quality datasets has largely hindered the advancement of AIM.
Emotions are complex.
Therefore, we seek to build EmoEditSet with diverse and representative examples for effective model training.
In Fig.~\ref{fig:method_1}, we represent each emotion with various representative semantic summaries (emotion attribution) and collect, generate and filter emotion data pairs (data construction).

\paragraph{Emotion Attribution}

EmoSet is a recently introduced large-scale visual emotion dataset~\cite{yang2023emoset}.
However, attribute labels within it are limited.
While some important emotion elements are missing (\eg, firework, ghost), some existing attributes show emotion ambiguity (\eg, rose, sunset).
Consequently, we conduct clustering on EmoSet to identify the common visual cues for each emotion in Fig.~\ref{fig:method_1} (a).
Given the significant correlation between emotions and semantics~\cite{brosch2010perception}, we experiment with semantic embeddings generated by CLIP~\cite{radford2021learning} and DINOv2~\cite{oquab2023dinov2}.
We observe that, for our purposes, CLIP captures visual semantics more effectively and eventually chose it for clustering.
Several post-processing steps are implemented to eliminate clusters characterized by a low number of images, excessive pixel-wise similarity, and low emotion score.
For more details, please refer to the supplementary material.

Following clustering and filtering, we designate the remaining $N$ clusters as \emph{factors} for each emotion.
We employ GPT-4V to assign a content summary to each factor and categorize the factors into four different types: object, scene, action, and facial expression. 
For instance, one of the factors shown in Fig.~\ref{fig:method_1} (a) is summarized as ``Books with flowers'', and classified into the ``Object'' category. 
The emotion factor tree is structured hierarchically, where each of the semantic summaries at the leaf nodes can evoke the emotion at the root node.
EmoSet comprises eight distinct emotions~\cite{mikels2005emotional}, for each of which a specific factor tree is constructed.
These emotions include \textit{amusement}, \textit{awe}, \textit{contentment}, \textit{excitement}, \textit{anger}, \textit{disgust}, \textit{fear}, and \textit{sadness}, with the first four as \textit{positive} and the last four as \textit{negative}.

\paragraph{Data Construction}

Aiming for larger data scale and greater image diversity, we collect images from multiple sources, including MagicBrush~\cite{zhang2024magicbrush}, MA5K~\cite{shi2021learning} and Unsplash\footnotemark[1].
While images in the first two sources are collected from social media, those from unsplash are more artistic.
We utilize IP2P~\cite{brooks2023instructpix2pix} to generate emotion data pairs, with instructions derived from the emotion factor trees.
Given the knowledge gap between GPT-4V and Stable Diffusion~\cite{rombach2022high}, we combine factors with high semantic similarity, and eliminate those with high content abstractness to enhance generation results.
Taking \textit{contentment} as an example, given an input image, IP2P receives instructions like ``Add colorful butterfly'', generating $N$ images simultaneously.
For more details, please refer to the supplementary material.
\footnotetext[1]{\url{https://unsplash.com/data}}

We then filter the $N$ target candidates through machine evaluation and human review.
As depicted in Fig.~\ref{fig:method_1} (b), we utilize CLIP image similarity~\cite{radford2021learning}, CLIP text similarity~\cite{radford2021learning}, Aesthetic score~\cite{schuhmann2022laion} and propose Emotion score as evaluation metrics.
To ensure appropriate pixel-wise consistency between the source and target images, we employ CLIP image similarity with a threshold ranging from 0.75 to 0.9.
Subsequently, CLIP text similarity is utilized to assess editing quality by comparing the text prompt with the target image, with a range of 0.25 to 1.
Since our task aims to evoke emotions through image manipulation, predicted score after softmax for the intended category, \ie, Emotion score, should exceed 0.3.
We introduce Aesthetic score to select the optimal target image, considering content consistency and visual appeal.
While these metrics greatly improve data quality, undesired edits still occur.
As shown in Fig.~\ref{fig:method_1} (b), some source-target pairs exhibit irrelevant structures, while certain target images are hard to interpret due to unreasonable or distorted content. 
Therefore, we incorporate manual review to further ensure the quality of the paired data.

Ultimately, EmoEditSet is constructed with a total of 40,120 image pairs, each formatted as a triplet: ``source-emotion-target''.
There are 15,531 source images and 40,120 data pairs, indicating that most images are modified to evoke multiple emotions, with an average of 2.6 directions per image.
An additional content instruction is provided for each image pair.
In Fig.~\ref{fig:method_1} (b), several content instructions can elicit \textit{contentment}.
Owing to the emotion factor tree, we can generate emotion-evoking targets with abundant semantic representations.

\subsection{EmoEdit}

To evoke specific emotions from viewers through image manipulation, we propose EmoEdit in Fig.~\ref{fig:method_2}.
Emotion adapter is trained using data from EmoEditSet, which can be directly plugged into existing diffusion-based methods to enhance their emotion knowledge at inference time.

\subsubsection{Emotion Adapter}

IP2P struggles to interpret and express emotions, as in Fig.~\ref{fig:teaser_1}, we thus aim to embed this capability into the model.
While directly fine-tuning IP2P with emotion data is a viable approach, it presents several drawbacks: (1) it is time-consuming and resource-intensive; (2) it lacks compatibility with models other than IP2P.
Recent studies have demonstrated that an \textit{adapter} can be seamlessly plugged into existing diffusion models without altering their original structure~\cite{zhang2023adding, ye2023ip, mou2024t2i}.
In this work, we investigate the feasibility of developing the Emotion adapter to make diffusion-based models emotion-aware.

AIM poses unique challenges compared to other editing tasks, since each emotion has various semantic representations, as shown in Fig.~\ref{fig:method_1}.
Consequently, automatically selecting the most appropriate representation for an input image remains a significant issue.
Q-Former~\cite{li2023blip} can leverage contextual information from one modality to improve the understanding in another.
We apply this capability to build the Emotion adapter, where target emotion and input image are encoded as $e_t$ and $e_i$ and then fused by attention mechanisms.
Specifically, the learned queries $q$ functions as an emotion dictionary, with $e_t$ and $e_i$ serving as indexes.
The self-attention mechanism first selects relevant semantics from this dictionary based on the target emotion (Eq.~\ref{eq:a_self}), while cross-attention identifies the most suitable representation by considering both the emotional context and the input image (Eq.~\ref{eq:a_cross}):
\begin{align}
	\label{eq:a_self}
	{{A}_{s}}\!=\!\textit{softmax} (\frac{[q;e_t]W_{q}^{s}{{\left( [q;e_t]W_{k}^{s} \right)}^{T}}}{\sqrt{{{d}_{k}}}})[q;e_t]W_{v}^{s},
\end{align}

\begin{align}
	\label{eq:a_cross}
	{{A}_{c}}\!=\!\textit{softmax} (\frac{{A}_{s}W_{q}^{c}{{\left( e_iW_{k}^{c} \right)}^{T}}}{\sqrt{{{d}_{k}}}})e_iW_{v}^{c},
\end{align}
where $W_{q}^{s}$, $W_{k}^{s}$, $W_{v}^{s}$ are the learned parameters in the self-attention block and $W_{q}^{c}$, $W_{k}^{c}$, $W_{v}^{c}$ the cross-attention block, $d_k$ is the dimension of keys.
Given emotion dictionary, target emotion and input image, our Emotion adapter integrates information from these three sources iteratively and generate the most appropriate emotion embedding $c_e$.

\subsubsection{Instruction Loss}

During training, as in Fig.~\ref{fig:method_2} (a), we optimize only the Emotion adapter while keeping the parameters in IP2P fixed.
Diffusion loss~\cite{rombach2022high} is introduced to train generative models by iteratively denoising random noise, facilitating high-quality sample generation and enhancing training stability.
Thus, we apply diffusion loss during the optimization process to guide the learning of pixel-level emotional representations in the paired data:
\begin{align}
	\label{eq:L_ldm}
	{{\mathcal{L}}_{LDM}}\!=\!{{\mathbb{E}}_{\mathcal E \left( x \right), {{c}_{i}}, {{c}_{e}},\epsilon ,t}}\!\left[ \left\| \epsilon -{{\epsilon }_{\theta }}\left( {{z}_{t}},t,\mathcal E \left( {{c}_{i}} \right),{{c}_{e}} \right) \right\|_{2}^{2} \right]\!,
\end{align}
where ${{c}_{i}}$ represents the input image, $\mathcal E $ denotes the latent encoder, $\epsilon$ refers to the added noise, ${{\epsilon }_{\theta }}$ indicates the denoising network and ${{z}_{t}}$ is the latent noise at time $t$.

By applying diffusion loss alone, however, EmoEdit overly emphasizes pixel-wise similarities, \ie, color and texture, resulting in undesired color artifacts in Fig.~\ref{fig:ablation}.
As in Fig.~\ref{fig:method_1} (a), each emotion can be evoked by several factors.
To capture the semantic variations in emotion data pairs, we introduce an instruction loss that guides the training process with explicit content instructions:
\begin{align}
	\label{eq:L_ins}
	{{\mathcal{L}}_{ins}}=\frac{1}{M}\left\| {{c}_{e}}- \mathcal E_{txt}\left( {{t}_{ins}} \right) \right\|_{2}^{2},
\end{align}
where ${{t}_{ins}}$ denotes the content instruction, $\mathcal E_{txt}$ indicates the text encoder and $M$ represents the number of elements in $c_e$.
By combining the two losses, EmoEdit is optimized with both pixel-level and semantic-level guidance.

\begin{figure*}
	\centering
	\includegraphics[width=\linewidth]{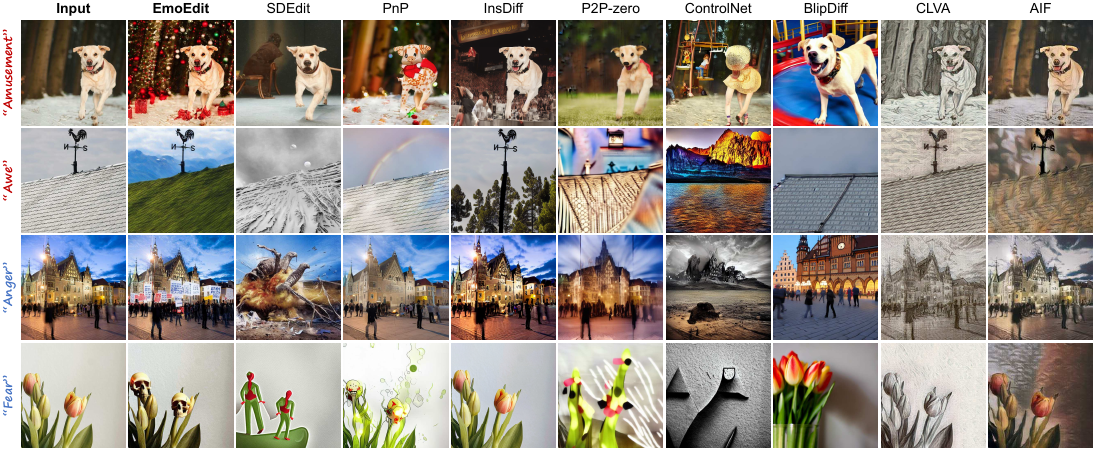}
	\vspace{-10pt}
	\caption{Comparison with the state-of-the-art methods, where EmoEdit surpasses others on emotion fidelity and structure integrity.
	}
	\label{fig:sota}
	\vspace{-10pt}
\end{figure*}

In the inference process, as shown in Fig.~\ref{fig:method_2} (b), given a user-provided image and a target emotion, EmoEdit generates an image that is emotionally faithful, contextually fit, and structurally preserved.
Notably, our Emotion adapter can be seamlessly plugged into other diffusion-based generative models to enhance their emotion knowledge beyond IP2P, as further validated in Fig.~\ref{fig:emotion adapter} and Fig.~\ref{fig:generation}.
\section{Experiments}
\label{sec:exp}
\begin{table}
	\centering
	\scriptsize
	\caption{Comparisons with the state-of-the-art methods on global editing, local editing and style-based AIM methods. 
		}
	\vspace{-5pt}
	\label{tab:exp_sota}
	\renewcommand\arraystretch{1}
	\setlength\tabcolsep{2.9pt}
	\begin{tabular}{l|cc|cc|cc}
		\toprule
		Method & PSNR $\uparrow$ & SSIM $\uparrow$ & LPIPS $\downarrow$  & CLIP-I $\uparrow$ & Emo-A $\uparrow$ & Emo-S $\uparrow$  \\
		\midrule
		SDEdit~\cite{meng2021sdedit} & \underline{15.43} & 0.415 & 0.459 &  0.638 & \underline{38.21\%} & \underline{0.221}  \\
		PnP~\cite{tumanyan2023plug}  & 14.41 & 0.436 & \underline{0.381} & \textbf{0.851} & 23.83\% & 0.095  \\
		\midrule
		InsDiff~\cite{geng2023instructdiffusion}  & 10.75 & 0.318 & 0.505 & 0.796 & 19.22\% & 0.060 \\
		P2P-Zero~\cite{parmar2023zero}  & 13.76 & 0.420 & 0.546 &  0.685  & 20.31\% & 0.067  \\	
		ControlNet~\cite{zhang2023adding}  & 11.98 & 0.292 & 0.603 &  0.686 & 36.33\% & 0.213 \\
		BlipDiff\cite{li2024blip}  & 9.00 & 0.249 & 0.654 &  0.810  & 18.06\% & 0.045  \\	
		\midrule
		CLVA~\cite{fu2022language} & 12.61 & 0.397 & 0.479 & 0.757 & 14.04\% & 0.017 \\
		AIF~\cite{weng2023affective} & 14.05 & \underline{0.537} & 0.493 & \underline{0.828} & 12.74\% & 0.004  \\
		\midrule
		EmoEdit &\textbf{16.62} & \textbf{0.571} &\textbf{0.289} & \underline{0.828} & \textbf{50.09\%} & \textbf{0.335} \\
		\bottomrule
	\end{tabular}
\vspace{-10pt}
\end{table}

\subsection{Dataset and Evaluation}

\paragraph{Dataset}

For fair comparisons, we compile an inference set of 405 images distinct from the training set, sourced from user uploads available online\footnotemark[1].
Each image is targeted with 8 emotion directions, resulting in a total of 3,240 image pairs.
To capture a broader spectrum of real data, our data comprises positive, negative and neutral images, categorized by a pre-trained emotion classifier on EmoSet~\cite{yang2023emoset}. 

\footnotetext[1]{\url{https://www.pexels.com/}}

\paragraph{Evaluation Metrics}

Given the multiple objectives of AIM, we assess our method based on three aspects: pixel-level (PSNR, SSIM), semantic-level (LPIPS, CLIP-I) and emotion-level (Emo-A, Emo-S).
PSNR measures the reconstruction quality by comparing the pixel-by-pixel similarity between the original and edited images, indicating the level of noise and distortion introduced during the manipulation process.
SSIM~\cite{wang2004image} evaluates the structural similarity of images by comparing low-level structural information, luminance, and contrast.
LPIPS~\cite{zhang2018unreasonable} leverages deep learning models to assess the perceptual similarity between images.
CLIP image similarity (CLIP-I)~\cite{radford2021learning} measures the semantic consistency between the original and edited images, ensuring the edits align with human perception and maintain contextual relevance.
Emotion accuracy (Emo-A)~\cite{yang2024emogen} assesses how well the edited image matches the targeted emotion, utilizing a pre-trained emotion classifier~\cite{yang2023emoset}.
AIM is a more challenging task than emotion generation because it requires both preserving structure and evoking emotions. 
Therefore, we also introduce a new metric to evaluate the increase in predicted scores for the desired emotion type, called the Emotion Incremental Score (Emo-S).
For more details, please refer to the supplementary material.

\begin{figure}
	\centering
	\includegraphics[width=\linewidth]{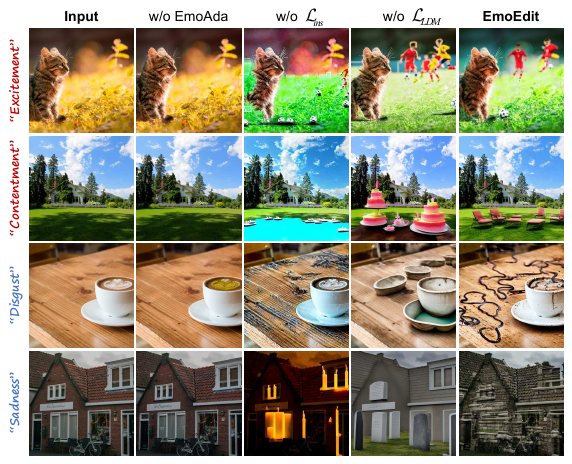}
	\vspace{-10pt}
	\caption{Ablation study on methodology. Emotion adapter, instruction loss and diffusion loss, are demonstrated to be vital.
	}
	\label{fig:ablation}
	\vspace{-10pt}
\end{figure}

\begin{figure*}
	\centering
	\includegraphics[width=\linewidth]{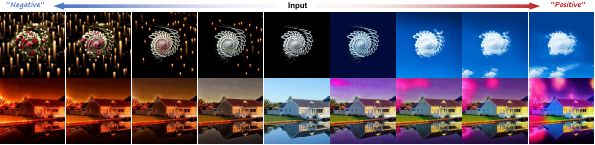}
	\vspace{-10pt}
	\caption{Ablation study on image guidance scale. EmoEdit can progressively edit an input image to different emotion polarities.
	}
	\label{fig:guidance-scale}
	\vspace{-10pt}
\end{figure*}

\subsection{Comparisons}

Since EmoEdit is the first attempt at content-aware editing in AIM, we compare our method with the relevant state-of-the-art techniques. 
These include global editing methods: SDEdit~\cite{meng2021sdedit}, PnP~\cite{tumanyan2023plug}, local editing methods: InsDiff~\cite{geng2023instructdiffusion}, P2P-zero~\cite{parmar2023zero}, ControlNet~\cite{zhang2023adding}, BlipDiff~\cite{li2024blip} and style-based AIM methods: CLVA~\cite{fu2022language}, AIF~\cite{weng2023affective}.

\begin{table}
	\centering
	\scriptsize
	\caption{User preference study. The numbers indicate the percentage of participants who vote for the result.}
	\vspace{-5pt}
	\label{tab:exp_userstudy}
	\renewcommand\arraystretch{1}
	\setlength\tabcolsep{5.5pt}
	\begin{tabular}{lcccccccc}
		\toprule
		Method & Structure integrity $\uparrow$  & Emotion fidelity $\uparrow$ & Balance $\uparrow$ \\
		\midrule
		SDEdit~\cite{meng2021sdedit} & 11.71$\pm$8.91\% & 10.85$\pm$7.50\% & 5.07$\pm$6.08\%  \\
		P2P-zero~\cite{parmar2023zero} & 3.05$\pm$3.60\% & 5.06$\pm$5.93\% & 0.94$\pm$3.05\%   \\
		BlipDiff~\cite{li2024blip}  & 15.12$\pm$16.13\% & 8.35$\pm$5.89\% & 4.88$\pm$10.45\%  \\
		EmoEdit  & \textbf{70.12$\pm$23.41\%}  & \textbf{75.73$\pm$16.44\%} & \textbf{89.12$\pm$14.56\%} \\
		\bottomrule
	\end{tabular}
\end{table}

\paragraph{Qualitative Comparisons}

We present the qualitative results in Fig.~\ref{fig:sota}. 
EmoEdit excels in both preserving structures and evoking emotions compared to other methods.
Most compared methods lack sufficient emotion knowledge, resulting in image distortion and severe artifacts.
Methods like SDEdit and ControlNet possess a certain level of emotional understanding, attempting to show people with \textit{awe} or \textit{anger}.
However, they struggle to choose contextually appropriate content for the given image.
Both CLVA and AIF are emotion style transfer methods that transform realistic images into artistic styles, but the modifications on different emotions are hard to distinguish.

\paragraph{Quantitative Comparisons}

In Table~\ref{tab:exp_sota}, EmoEdit outperforms other methods across various metrics. 
Due to a lack of emotion knowledge, most methods perform poorly on emotion-level metrics, with only 38.21\% versus 50.09\% in Emo-A, and 0.221 versus 0.335 in Emo-S. 
One of the biggest challenges in AIM is balancing emotion fidelity with structure integrity. 
However, aside from emotion-level metrics, EmoEdit also achieves the best results in pixel-level metrics like PSNR and SSIM. 
For CLIP-I, EmoEdit ranks second, slightly behind PnP, likely because content changes can decrease semantic similarity.
Among all compared methods, global editing techniques SDEdit and PnP perform relatively well, possibly because emotion effects are more global rather than local. 
CLVA and AIF aim to evoke emotions through artistic styles, which effectively preserves image structure but are restricted to elicit emotions in realistic images.

\begin{figure}
	\centering
	\includegraphics[width=\linewidth]{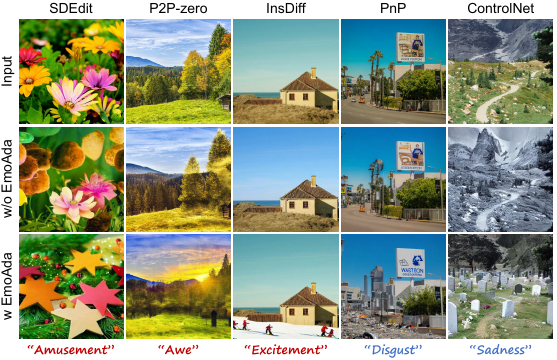}
	\vspace{-10pt}
	\caption{Emotion adapter can be effectively plugged into existing editing models to enrich their emotional knowledge.
	}
	\label{fig:emotion adapter}
	\vspace{-10pt}
\end{figure}

\begin{figure*}
	\centering
	\includegraphics[width=\linewidth]{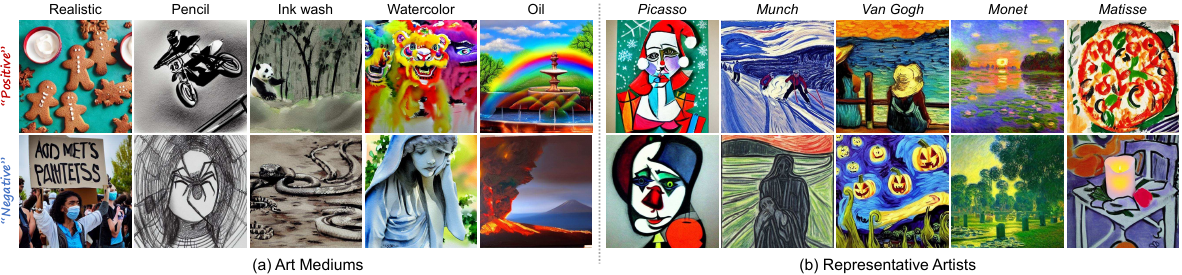}
	\vspace{-10pt}
	\caption{Emotion adapter can be extended to stylized image generation, preserving style and evoking emotions with clear semantics.
	}
	\label{fig:generation}
	\vspace{-10pt}
\end{figure*}

\paragraph{User Study}

We conducted a user study to assess whether humans prefer our method. 
We invited 41 participants of various ages, with each session lasting about 15 minutes.
The study included 40 sets of images, each featuring an original image alongside four edited versions from different methods: SDEdit, P2P-zero, BlipDiffusion, and EmoEdit.
Participants were shown a set of images and asked three questions: (1) Which image best preserves the structure? (2) Which image most strongly evokes the targeted emotion? (3) Which image achieves the best balance between structure and emotion?
Participants could choose one out of the four options and we calculate the vote percentage for each question.
Results in Table~\ref{tab:exp_userstudy} show that EmoEdit is the most preferred choice in all questions.
Despite the challenge of maintaining structure while conveying emotions, EmoEdit received the highest votes for both aspects, \ie, 70.12\% and 75.73\%. 
In terms of balancing both aspects, EmoEdit shows a clear advantage with 89.12\% support, confirming alignment with human perception.

\subsection{Ablation Study}
\paragraph{Methodology}

In Fig.~\ref{fig:ablation}, we evaluate the effectiveness of several key designs in EmoEdit, including Emotion adapter, instruction loss and diffusion loss.
Without the Emotion adapter (w/o EmoAda), images remain nearly identical, highlighting its significance.
Both diffusion loss and instruction loss are crucial for evoking emotions.
While diffusion loss well preserves the original structure, instruction loss enhances semantic clarity.
For instance, in the case of \textit{contentment}, using diffusion loss adds water to the grass without clear semantics, while using instruction loss adds cakes to the grass, without structural preservation.
EmoEdit adds lounge chairs, showcasing structure integrity, semantic clarity and contextual fitting.

\paragraph{Guidance Scale}

We show EmoEdit's editing results with variations in image guidance scale in Fig.~\ref{fig:guidance-scale}.
The middle image represents the input, while the left and right images illustrate different guidance scales for two emotion polarities: positive and negative.
We observe that as the image guidance scale decreases, emotion intensity increases while structure preservation diminishes.
Although evoking emotion and preserving structure are often contradictory, our method effectively balances them, as demonstrated in Table~\ref{tab:exp_sota} and Table~\ref{tab:exp_userstudy}. 
Users can customize the level of manipulation by adjusting guidance scale to suit their preferences.

\subsection{Applications}
Once EmoEdit is trained, Emotion adapter can be directly plugged into various diffusion-based models to enhance their emotion awareness with diverse semantics, covering both image editing task and image generation task. 

\paragraph{Emotion-enhanced Editing Models}
As most existing editing models in Fig.~\ref{fig:sota} lack emotion knowledge, we experiment to investigate whether Emotion adapter can augment their emotion capabilities in a plug-and-play manner.
In Fig.~\ref{fig:emotion adapter}, when Emotion adapter is plugged to the original method (w EmoAda), these models are capable of generating images with emotion fidelity and contextual fit.
For example, ControlNet transforms the input image into black-and-white, while additional elements, such as thumb stones, are introduced after attaching the Emotion adapter.  

\paragraph{Emotion-aware Stylized Image Generation}
Apart from editing task, Emotion adapter can also be extended to stylized image generation task in Fig.~\ref{fig:generation}.
Once trained, Emotion adapter functions as an emotional interpreter, encoding each emotion polarity (\ie, positive, negative) into distinct semantic representations.
We combine the Emotion adapter with Composable Diffusion~\cite{liu2022compositional}, using 5 art mediums and 5 representative artists as style prompts.
Results show that each generated images preserve art styles while evoke specified emotions with clear semantics.
Taking \textit{Monet} as an example, the sunset evokes \textit{awe (positive)}, while the graveyard makes people feel \textit{sadness (negative)}.

\section{Conclusion}
\label{sec:conclusion}

\paragraph{Discussion}
We present EmoEdit, a method designed to evoke emotions by modifying the content of user-provided images.
EmoEditSet is first constructed with 40,120 image pairs, serving as a data foundation for AIM.
Emotion adapter is proposed and trained using both diffusion loss and the designed instruction loss, which can be directly plugged into diffusion-based generative models to enhance their emotion-awareness. 
Our method is evaluated both qualitatively and quantitatively, demonstrating a strong balance between structure preservation and emotion faithfulness. 
User study and ablation study validate the effectiveness of EmoEdit, while applications further highlight the portability of Emotion adapter and the rich emotion knowledge embedded within it.

\paragraph{Limitations}
In the real world, there are numerous visual emotion factors beyond those in the constructed emotion factor trees.
Besides, due to the inherent complexity of emotions, more than eight emotion categories exist. 
Our method is heavily dependent on EmoSet, suggesting that the limited data may introduce potential biases and constraints.
Expanding the range of emotion factors and categories would improve the quality and diversity of the editing results.
Since AIM is a highly human-centered task, future work should place greater emphasis on human interaction.
Evaluation metrics, such as emotion accuracy, are largely dependent on the training data and are prone to bias, requiring more human feedback. 
Additionally, increased flexibility would allow users to tailor the editing process to suit their personal tastes and preferences.


{
	\small
	\bibliographystyle{ieeenat_fullname}
	\bibliography{EmoEdit}
}


\end{document}